\documentclass{ifacconf}
\usepackage[utf8]{inputenc}
\usepackage{amsmath}

\usepackage{amsfonts}
\usepackage{amssymb}
\usepackage{graphicx}
\usepackage{natbib}        

\usepackage{mathtools}
\usepackage{bm}
\usepackage{tabularx}

\usepackage{xcolor}

\begin{document}
\begin{frontmatter}
\title{Energy Tank-based Control Framework for Satisfying the ISO/TS 15066 Constraint}

\author[First]{F. Benzi} 
\author[First]{F. Ferraguti} 
\author[First]{C. Secchi}

\address[First]{Department of Sciences and Methods of Engineering,
     University of Modena and Reggio Emilia, Italy (e-mail: federico.benzi@unimore.it, federica.ferraguti@unimore.it, cristian.secchi@unimore.it).}

\begin{abstract}
    The technical specification ISO/TS 15066 provides the foundational elements for assessing the safety of collaborative human-robot cells, which are the cornerstone of the modern industrial paradigm. The standard implementation of the ISO/TS 15066 procedure, however, often results in conservative motions of the robot, with consequently low performance of the cell. In this paper, we propose an energy tank-based approach that allows to directly satisfy the energetic bounds imposed by the ISO/TS 15066, thus avoiding the introduction of conservative modeling and assumptions. The proposed approach has been successfully validated in simulation.
\end{abstract}

\end{frontmatter}
\section{Introduction} \label{sec: intro}
Collaborative robotics is a fast growing field in the industrial setting and it allows direct robot and operator interface without traditional safeguarding. As a consequence of the introduction of human-robot collaboration technologies, great importance has been attributed to robot safety standards, which have been updated to address new co-working scenarios. In particular, the international \cite{ISO_10218-1:2011} and \cite{ISO_10218-2:2011} safety standards have identified specific applications and criteria where collaborative operations can occur and they have been integrated by the introduction of the technical specification \cite{ISO_TS_15066:2016}. The safety regulations identify four \textit{collaborative operations} which can be adopted depending on the requirement of the application concerned and the design of the robot system: Safety-rated Monitored Stop (SMS), Speed and Separation Monitoring (SSM), Power and Force Limiting (PFL) and Hand Guiding (HG). These modalities can be applied in combination in order to achieve higher levels of productivity, while still preserving safety of the human operators \cite{pupa2021}. However, in the industrial practice, PFL is typically adopted for tasks in which intentional or incidental contact between the human and the robot can occur. Safety is guaranteed by limiting the power and the force during the contact to values at which the risk of injuries is not expected and by imposing speed limits that guarantee safe human-robot contacts. 
Current approaches for implementing the ISO/TS 15066 guidelines lead to a conservative behavior (e.g. low velocity) of the robot and, consequently, to poor performance of the collaborative cell. In \cite{ferraguti2020control}, Control Barrier Functions (CBF) \cite{Ferraguti2022RAM} have been exploited for enhancing the performance of a robot operating in a collaborative cell while satisfying the PFL velocity limit. However, the definition of the safe set is very conservative since it is a collection of ellipses and the overall CBF is non-smooth due to the abrupt activation and deactivation of the ellipsoids.
Moreover, as stated in the regulation itself, the velocity limit computed according to the ISO/TS 15066 is based on conservative assumptions and this results in robot motions that are still very slow. Indeed, the original formulation of the PFL constraint is a limitation onto the exchanged energy, which is then turned into a velocity constraint under strong assumptions (i.e., the assumption of a fully inelastic contact, a two-body model after the impact). Recently, an energy-based approach has been proposed in \cite{lachner2021energy} to reduce the safety-related parameters to be determined in physical Human-Robot Interaction (pHRI) applications to a single energy value. However, the proposed architecture relies heavily on the precise knowledge of the dynamic model of the manipulator and, in particular, of its inertia matrix, which is not always available or known with sufficient accuracy. 

To overcome these issues, in this paper we leverage energy tanks and passivity-based control in order to directly address the PFL energetic constraint. To this aim, we exploit the modulated energy-tanks presented in recent works \cite{benzi2021optimal,benzi2022shared}. These are energy storing elements which have proven effective in ensuring safety layers for collaborative applications, by passively implementing any desired dynamics. By simply controlling the power flow in the system, these techniques can, in fact, bound the energy of the interconnected system without requiring any knowledge of its model, while at the same time providing formal guarantees of robust stability.

Thus, the contribution of this paper is an energy tank-based control architecture designed to comply directly with the time-varying energetic bounds of the \cite{ISO_TS_15066:2016}. This allows us to overcome the modeling assumptions underlying the conservative velocity limitation. At the same time, owing to the passivity-based nature of the formulation, our approach does not require the precise knowledge of the dynamic model of the robot, improving upon this limitation of \cite{lachner2021energy}.

\section{Background on Power and Force Limiting}\label{sec:background}
As specified in ISO/TS 15066, PFL is a form of collaborative operation in which intentional or incidental contact between the moving robot and a human body region can occur. Safety is achieved by limiting the contact pressure and force to values at which injuries and risks are not to be expected. These maximum values (bio-mechanical load limits) have been established on the basis of pain sensitivity thresholds. The Annex A of the ISO/TS 15066 reports the maximum contact pressure and force for each body area. From these values, the maximum energy transfer due to an hypothetical human-robot contact is determined, treating the body part as a linear spring. 
Formally, for each body region the maximum permissible energy transfer can be calculated as \cite{ISO_TS_15066:2016}:
\begin{equation}\label{eq:max_energy}
    E_{max} = \frac{f^2_{max}}{2k},
\end{equation}
where $f_{max} \in \mathbb{R}$ is the maximum contact force for the body region and $k \in \mathbb{R}$ is the related effective spring constant.  

The energy transfer limit established in \eqref{eq:max_energy} is then used to compute the maximum velocity at which the robot can move into the collaborative workspace, while ensuring safety in case of collision. Let us define $v_{rel} \in \mathbb{R}$ as the relative velocity between the robot and the human along the minimum distance direction between the threatened body part and the closest link of the robot.

In order to derive a relationship between the relative velocity and the contact force, the ISO/TS 15066 assumes a fully inelastic contact during the collision, in which the total kinetic energy of the two-body system is transferred to the human body part. The resulting balance is: 
\begin{equation}\label{eq:max_vel_tran}
E = \frac{f^2}{2k} = \frac{1}{2} \mu v^2_{rel},
\end{equation}
where $E \in \mathbb{R}$ is the energy transferred and $\mu \in \mathbb{R}$ is the reduced mass of the two body system, computed as:
\begin{equation}\label{eq:reduced_mass}
\mu = (m_h^{-1} + m_r^{-1})^{-1},
\end{equation}
in which $m_h \in \mathbb{R}$ is the mass of the human body area and $m_r \in \mathbb{R}$ the mass of the robot, the latter computed as:
\begin{equation} \label{eq:robot_mass_iso}
m_r = (M / 2) + m_L,
\end{equation}
being $M \in \mathbb{R}$ the mass of the moving parts of the robot, while $m_L \in \mathbb{R}$ is the total payload. Both $m_{h}$ and $k$ are tabulated in the Annex A of the ISO/TS 15066, for each body region. Thus, combining \eqref{eq:max_energy} and \eqref{eq:max_vel_tran}, safety in PFL operations is encoded by the following constraint on the velocity between the robot and a given human body part:
\begin{equation}\label{eq:vISO}
    v_{rel} \leq v_{max}=\frac{f_{max}}{\sqrt{\mu k}}.
\end{equation}

\section{Problem Statement}\label{sec: problem}
The velocity limit computed according to the ISO/TS 15066 and given by \eqref{eq:vISO} is severely conservative, due to the assumptions underlying its formulation. In particular, we could list the following drawbacks related to the direct implementation of the ISO/TS 15066: 
\begin{enumerate}
    \item As showcased, e.g., in \cite{khatib1995inertial} and \cite{haddadin2010safe}, the force exerted during a contact scenario $f$ depends on both the robot configuration and the direction of the contact. Therefore, the computation of the effective robot mass $m_r$ in \eqref{eq:robot_mass_iso} is often inaccurate and can result in large over-estimations.
    \item The assumption that the transient contact between a robot and a human body part results in a fully inelastic two-body collision is highly conservative. The actual transient contact scenario is, in most cases, an intermediate condition between a fully elastic and a fully inelastic collision. Thus, the value of $v_{max}$ computed through the model in \eqref{eq:vISO} would end up being in most cases inaccurate and/or conservative.
    \item Even if the ISO/TS 15066 limits the energy exchange, there is no formal analysis related to the stability of the system while in contact with the human, as well as while switching from interaction to free motion and vice-versa. Thus, instabilities might arise during the contact, threatening the safety of the human.
\end{enumerate}
Points 1) and 2) in the previous list highlight that using the velocity limit \eqref{eq:vISO} can be severely conservative. In the remainder of this section we discuss how, by applying the robot mass evaluation proposed by the ISO/TS 15066 through \eqref{eq:robot_mass_iso}, we obtain in most cases a conservative value. 

Let us consider a torque-controlled fully actuated $n-$DOF manipulator represented by the following Euler-Lagrange dynamic model:
\begin{equation} \label{eq: robot_dyn_model}
\textbf{M}(\textbf{q})\Ddot{\textbf{q}}(t) + \textbf{C}(\textbf{q}, \dot{\textbf{q}})\dot{\textbf{q}}(t) + \textbf{g}(\textbf{q}) =\boldsymbol\tau(t)+\textbf{J(q)}^T \mathbf{F}_{e}(t),
\end{equation}
where $\textbf{q}(t) \in \mathbb{R}^n$ is the vector of joint variables, $\textbf{M}(\textbf{q}) \in \mathbb{R}^{n \times n}$ is the positive definite and symmetric inertia matrix, $\textbf{C}(\textbf{q}, \dot{\textbf{q}}) \in \mathbb{R}^{n \times n}$ encompasses Coriolis and centrifugal effects and $\textbf{g}(\textbf{q}) \in \mathbb{R}^{n}$ is the gravitational term. The vector  $\boldsymbol\tau(t)\in \mathbb{R}^{n}$ represents
the controlled joint torques and the term $\textbf{J(q)}^T \mathbf{F}_{e}(t)\in \mathbb{R}^{n}$ represents the torque applied to the joints because of the
external wrench $\mathbf{F}_{e}(t)\in \mathbb{R}^{m}$ applied on the end-effector.
Let us assume that the robot is  carrying a negligible load, i.e. $m_L \approx 0$ in \eqref{eq:robot_mass_iso}. Thus, according to the ISO/TS 15066, the effective mass of the robot can be computed as $m_r = \frac{M}{2}$. The effective mass value $m_r$ is then used in \eqref{eq:max_vel_tran}, through $\mu$, to evaluate the velocity $v_{rel}$. However, the value of the contact force $f$ along a given direction, indicated by the unit vector $\textbf{n} \in \mathbb{R}^3$, depends upon the \textit{apparent mass} $m_{app} \in \mathbb{R}$ of the robot at the contact point \cite{khatib1995inertial, haddadin2010safe}, namely
\begin{equation} \label{eq: apparent_mass}
m_{app} = \left(\textbf{n}^T \bm{\Lambda}(\textbf{q})^{-1} \textbf{n}\right)^{-1},
\end{equation}
in which $\bm{\Lambda}(\textbf{q})^{-1} \in \mathbb{R}^{3 \times 3}$ is the end-point mobility tensor \cite{hogan1984impedance}. For a manipulator, this measure provides a description of the inertial properties at a given configuration and can be computed, considering a pure translational motion, as
\begin{equation}\label{eq: end_point_tensor}
\bm{\Lambda}(\textbf{q})^{-1} = \textbf{J}_v(\textbf{q})\textbf{M}(\textbf{q})^{-1}\textbf{J}_v(\textbf{q})^{T},
\end{equation}
where $\textbf{J}_v(\textbf{q}) \in \mathbb{R}^{3 \times n}$ is the Jacobian associated with the linear velocity at the operational point.

Since the robot mass in \eqref{eq:robot_mass_iso} is computed without information regarding the configuration of the robot and the direction of contact, the estimated value could be significantly larger than $m_{app}$, as thoroughly showcased in \cite{haddadin2009requirements}. This, in turn, leads to smaller values of $v_{max}$ according to \eqref{eq:vISO}, producing an over-conservative velocity limit for the robot. A thorough comparison in this regard is presented in \cite{kirschner2021notion}.

A possible solution would be to directly use the apparent mass in \eqref{eq:robot_mass_iso}, instead of $m_r$. However, the computation of $m_{app}$ requires the precise knowledge of the dynamic model of the robot, specifically of the inertia matrix $\textbf{M}(\textbf{q})$, which is not always available or known with sufficient accuracy.

Therefore, in this paper we propose a control architecture which exploits directly the energetic constraint in \eqref{eq:max_energy} for satisfying the safety requirements of PFL in the ISO/TS 15066, via an energy tank-based formulation. In this way, we can avoid  the conservative assumptions underlying the derivation of the velocity limit in \eqref{eq:vISO}, thus overcoming the drawbacks presented in points 1) and 2). Moreover, we provide a model-agnostic passivity-based framework in order to guarantee a robustly stable behavior of the overall system both in free motion and in contact with the human, thus addressing the issue described in point 3). 

\section{Optimized Tank-based Control Architecture for Ensuring Variable Bounds on the Kinetic Energy}\label{sec: bounds}

Consider a robot that needs to execute a task in a collaborative cell.  Let \mbox{$0<H_1\leq H_2 \leq \cdots \leq H_N $} be the bounds on the energy transfer imposed by PFL in \eqref{eq:max_energy} for guaranteeing safety during contact with each of the $N$ human body parts. Since the closest body part can change online, the bound to guarantee is time-varying.

Since the nature of the problem is energetic, we provide an energy-based solution exploiting the modulated energy tank \cite{benzi2022shared, benzi2021optimal}. In this section, we first provide a short background on modulated energy tanks. Then, we describe the control architecture to ensure a single bound on the kinetic energy, both in free motion and in contact phase. Finally, we extend it for switching among the different bounds imposed by PFL.

\subsection{Background on Modulated Energy Tanks}\label{sec:modulated}
The energy tank is an energy storing element, traditionally employed for storing the energy dissipated by the controlled system. It can be generally represented as
\begin{equation}
    \label{eq: tankeqs}
    \begin{cases}
    \dot{x}_t(t) = u_t(t) \\
    y_t(t) = \frac{\partial T(t)}{\partial x_t(t)} = x_t(t),
    \end{cases}
\end{equation}
where $x_{t}(t) \in \mathbb{R}$ is the tank state, \mbox{$(u_t(t), y_t(t)) \in \mathbb{R} \times \mathbb{R}$} is the power port of the tank, with associated energy function 
\begin{equation} \label{eq: tank_energy}
    T(x_{t}(t)) = \frac{1}{2} x_{t}^2(t).    
\end{equation}
The tank stores/releases energy by means of the port $(u_t(t), y_t(t))$. In particular, this can be exploited for reproducing any desired port-behavior (\cite{benzi2022shared}):
\begin{equation}\label{eq: modulated_io}
    \begin{cases} 
        u_t(t) = \textbf{a}(t)^T \textbf{u}(t) \\
        \textbf{y}(t) = \textbf{a}(t) y_t(t),
    \end{cases}
\end{equation}
where $(\textbf{u}(t), \textbf{y}(t)) \in \mathbb{R}^n \times \mathbb{R}^n$ is the I-O port of the system and $\textbf{a}(t) \in \mathbb{R}^n$ is a modulating term, defined as
\begin{equation} \label{eq: modulating_term}
    \textbf{a}(t) = \frac{\bm{\gamma}(t)}{x_t(t)},
\end{equation}
with $\bm{\gamma}(t) \in \mathbb{R}^n$ being the desired value for the output $\textbf{y}(t)$. By embedding \eqref{eq: modulated_io} in \eqref{eq: tankeqs} we obtain
\begin{equation}\label{eq: modulated_tank}
    \begin{cases} 
        \dot{x}_t(t) = \textbf{a}(t)^T \textbf{u}(t) \\
        \textbf{y}(t) = \textbf{a}(t) y_t(t) = \bm{\gamma}(t),
    \end{cases}
\end{equation}
which shows how the modulating term $\textbf{a}(t)$ allows to control the energy flowing in the interconnected system to obtain any desired value for the port output $\textbf{y}(t)$.
A lower bound $\varepsilon > 0$ such that $T(x_t(t)) \geq \varepsilon \quad \forall t \geq 0$ must be enforced, in order to avoid numeric singularities in \eqref{eq: modulating_term}. Additionally, an upper tank energy bound has to be set, or practically unstable behaviors could be implemented.

The tank \eqref{eq: modulated_tank} has been proven to implement a passive exchange of energy as long as the tank is not empty (\cite{benzi2022shared}), thus the following proposition holds. 
\begin{prop}[\textbf{\cite{benzi2022shared}, Prop. 1}] \label{prop: passivity_modulated_tank}
    If \\$T(x_t(t)) \geq \varepsilon $ $\forall t \geq 0$, then the
    modulated tank \eqref{eq: modulated_tank} is passive independently of the desired value of $\bm{\gamma}(t)$.
\end{prop}
Thus, as long as some energy is present in the tank, any desired port behavior can be passively implemented.

\subsection{Tank-based Control Architecture for Energy Limitation}\label{sec:single}
\begin{figure}[t]
	\centering
	\includegraphics[width= 0.9\columnwidth]{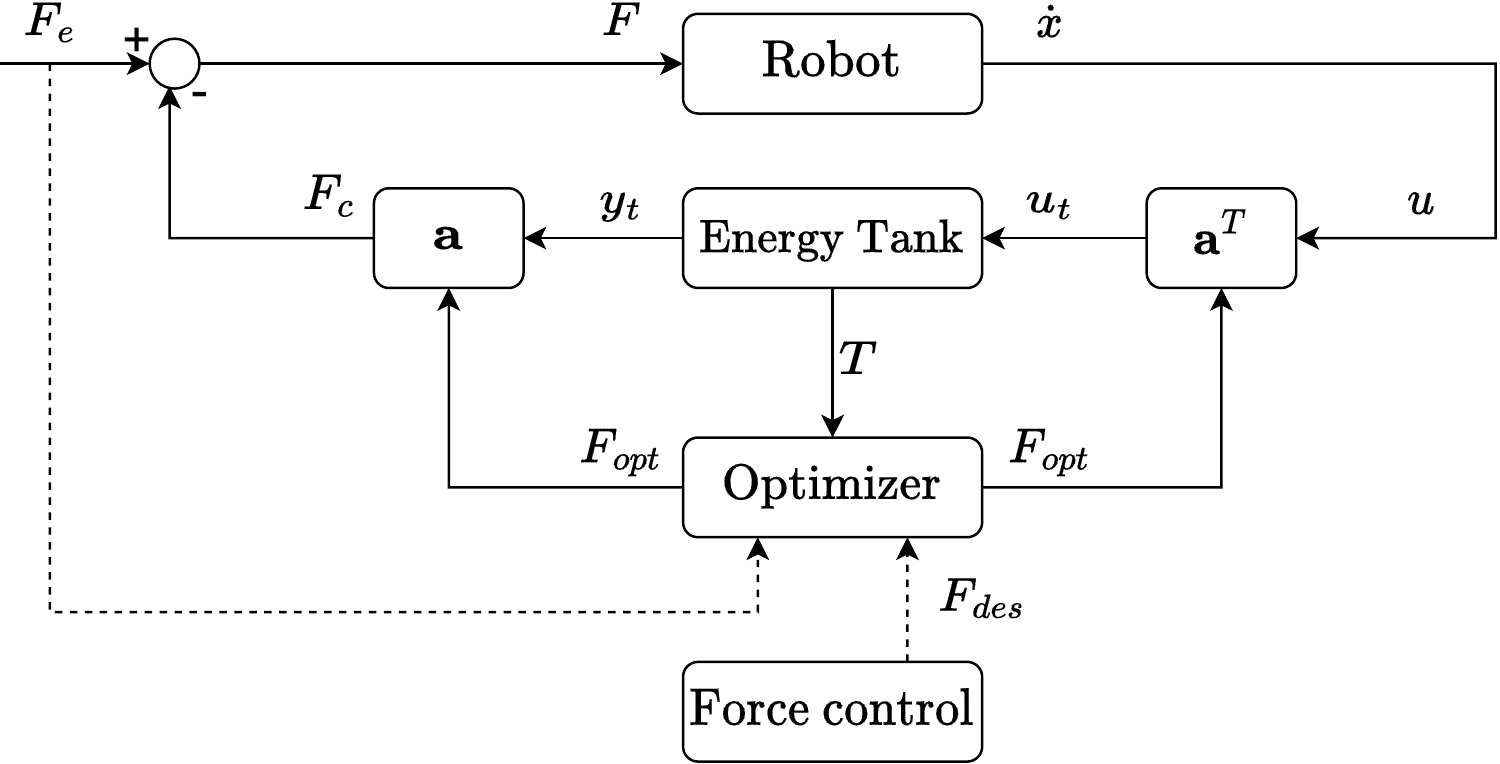}
	\caption{Modulated energy tank control architecture. The optimizer provides the closest passive approximation $\mathbf{F}_{opt}$, which is then used to properly modulate, via the term $\textbf{a}(t)$, the power flow at the ports of the tank.}
	\label{fig:ATA_no_ext}
\end{figure}

Consider the model of the robot \eqref{eq: robot_dyn_model} in operational space:
\begin{equation}\label{eq: cart_dyn_model}
    \bm{\Lambda}(\textbf{x})\Ddot{\textbf{x}}(t) + \textbf{S}(\textbf{x}, \dot{\textbf{x}})\dot{\textbf{x}}(t)= \textbf{F}(t),  
\end{equation}
 where $\bm{\Lambda}(\textbf{x}) \in \mathbb{R}^{m \times m}$ and $\textbf{S}(\textbf{x}, \dot{\textbf{x}}) \in \mathbb{R}^{m \times m}$ are, respectively, the inertia matrix and the centrifugal and Coriolis matrix in the operational space, with $\textbf{x}(t) \in \mathbb{R}^{m}$ being the Cartesian pose (position and orientation) of the end-effector and $\textbf{F}(t) \in \mathbb{R}^m$ the operational force vector. We decompose the latter into two terms, 
\begin{equation}\label{eq: force_decomposed}
    \mathbf{F}(t) = -\mathbf{F}_c(t) + \mathbf{F}_{e}(t),
\end{equation}
with $\textbf{F}_c(t) \in \mathbb{R}^m$ being the control forces due to the actuation and $\mathbf{F}_{e}(t) \in \mathbb{R}^m$ being the external wrench, due to the interaction with the environment or the human.
We can then compute the kinetic energy of \eqref{eq: cart_dyn_model} as 
\begin{equation}
 H(t)=\frac{1}{2} \dot{\textbf{x}}(t)^T \bm{\Lambda}(\textbf{x}) \dot{\textbf{x}}(t) \geq 0   
\end{equation}
It is well known that \eqref{eq: cart_dyn_model} is passive with respect to the pair $(\textbf{F}(t), \dot{\textbf{x}}(t))$, since \mbox{$\textbf{F}(t)^T\dot{\textbf{x}}(t)=\dot{H}(t)$} (see, e.g., \cite{secchi2007control}). Our initial goal is to limit $H(t)$ to a fixed value $\bar H\in \mathbb{R}$ during free motion (i.e., $\mathbf{F}_{e}(t)= \mathbf{0}$). Formally, we aim at bounding $H(t)\leq \bar H$ with $\bar H \geq H(0)$. This can be accomplished by interconnecting \eqref{eq: cart_dyn_model} with the modulated energy tank \eqref{eq: modulated_tank} as shown in Fig.~\ref{fig:ATA_no_ext}. 
Here, the input and the output of the modulated tank are the Cartesian velocity $\dot{\bm{x}}(t)\in \mathbb{R}^m$ and the control forces $\mathbf{F}_{c}(t)\in \mathbb{R}^m$, respectively, while the goal is to implement the desired control action $\mathbf{F}_{des}(t)\in \mathbb{R}^m$. The control input is then provided to an optimization block, which computes the best passive approximation $\mathbf{F}_{opt}(t)\in \mathbb{R}^m$ of the control action $\mathbf{F}_{des}(t)$ considering the energy available in the tank and the desired behavior. The output $\mathbf{F}_{opt}(t)$ is then used for modulating the energy flowing into/out of the energy tank according to \eqref{eq: modulated_tank} by setting $\bm{\gamma}(t)=\mathbf{F}_{opt}(t)$.

Since \eqref{eq: cart_dyn_model} is passive and the negative feedback interconnection is power preserving, the passivity of \eqref{eq: modulated_tank} would guarantee the passivity of the overall controlled system and, consequently, a stable behavior both during free motion and interaction \cite{secchi2007control}.  According to Prop. \ref{prop: passivity_modulated_tank}, passivity of \eqref{eq: modulated_tank} is guaranteed if $T(x_t(t)) \geq \varepsilon $. Thus, given the desired input $\mathbf{F}_{des}(t)$, it is possible to find $\mathbf{F}_{opt}(t)$ by solving the following optimization problem:
\begin{equation} \label{eq: start}
    \begin{aligned}
        & \underset{\mathbf{F}_{c}(t)}{\text{minimize}}
        & & ||\mathbf{F}_{c}(t) - \mathbf{F}_{des}(t)||^2 \\
        & \text{subject to}
        & & T(x_t(t)) \geq \varepsilon.
        \end{aligned}
\end{equation}
The optimization problem \eqref{eq: start} can be rewritten for a discrete time implementation, $\tau>0$ being the cycle time, as shown in \cite{benzi2022shared}, to obtain a convex optimization problem, suitable for real-time implementations:
\begin{equation} \label{eq: discrete_start}
    \begin{aligned}
        & \underset{\mathbf{F}_{c}(k)}{\text{minimize}}
        & & ||\mathbf{F}_{c}(k) - \mathbf{F}_{des}(k)||^2 \\
        & \text{subject to}
        & & \tau {\underbrace{\vphantom{\sum_{i=0}^{k-1}\tau \textbf{F}_c(i)^T\dot{\textbf{x}}(i)}\textbf{F}_c(k)^T\dot{\textbf{x}}(k)}_{\dot{T}(k)}}+{\underbrace{\sum_{i=0}^{k-1}\tau \textbf{F}_c(i)^T\dot{\textbf{x}}(i)}_{T(k-1)}} \geq \varepsilon.
        \end{aligned}
\end{equation}
Notice how, deploying this architecture, we only allow a set amount of energy to the robot for implementing the task in free motion, i.e. its kinetic energy will at most be equal to $H(t) = H(0) + T(0) - \varepsilon$. Thus, by a proper choice of $T(0)$, we can guarantee a fixed bound $\bar H$ on $H(t)$:
\begin{prop}\label{prop:ATAnoF}
Consider the robot \eqref{eq: cart_dyn_model} interconnected with the tank \eqref{eq: modulated_tank} as in Fig.~\ref{fig:ATA_no_ext}. If $T(0)=\bar H-H(0)+\varepsilon$ and $T(t) \geq \varepsilon$ $\forall t \geq 0$, then $H(t)\leq \bar H\,\, \forall t\geq 0$ in free motion ($\mathbf{F}_{e}(t)= \mathbf{0}$).
\end{prop}
\begin{pf}
Since 
\begin{equation}
  \label{eq:powpres}
  \left\{
    \begin{array}[l]{ll}
    \mathbf{u}(t)=\mathbf{\dot x}(t) \\
    \mathbf{y}(t)=\mathbf{F}_{c}(t)=-\mathbf{F}(t)    
    \end{array}
    \right.
\end{equation}
and 
\begin{equation}
    \mathbf{F}(t)^T\mathbf{\dot x}(t)=-\mathbf{u}(t)^T\mathbf{y}(t),
\end{equation}
we have that
\begin{equation}
  \label{eq:ATAnoF-balance}
  \dot{H}(t)+\dot{T}(t)=0,
\end{equation}
whence
\begin{equation}
  \label{eq:ATAnoF-energy}
  H(t)=H(0)+T(0)-T(t)\leq H(0)+T(0)-\varepsilon,
\end{equation}
in which the inequality comes from the assumption that \mbox{$T(t)\geq \varepsilon$ $\forall t\geq0$}. Then, by setting \mbox{$T(0) = \bar H - H(0) + \varepsilon$}, we get \mbox{$H(t)\leq \bar H$}, thus concluding the proof.
\end{pf}

The condition on the lower bound can be enforced by synthesizing the control input $\mathbf{F}_{c}(t)$ via the optimization problem \eqref{eq: discrete_start}. We can thus compute the best approximation of $\mathbf{F}_{des}(t)$ complying with the energetic limitation. 

The approximated input $\mathbf{F}_{c}$ in \eqref{eq: discrete_start}, however, does not necessarily preserve the direction of the desired behavior $\mathbf{F}_{des}$. As, in this work, we focus on safely accomplishing industrial tasks, in which maintaining a given direction can be critical, we further modify the formulation as:
\begin{equation} \label{eq: opt_max_energy_freemotion}
    \begin{aligned}
        & \underset{\alpha(k)}{\text{minimize}}
        & & ||\alpha \mathbf{F}_{des}(k) - \mathbf{F}_{des}(k)||^2 \\
        & \text{subject to}
        & & \tau \alpha \textbf{F}_{des}(k)^T\dot{\textbf{x}}(k)+ 
        T(k-1)\geq \varepsilon.
        \end{aligned}
\end{equation}
Thus, by setting $\mathbf{F}_{c} = \alpha \mathbf{F}_{des}$, we obtain the safe version of $\mathbf{F}_{des}$ preserving the intended direction. As $\alpha = 0$ is always admissible, the problem always possesses a solution. 

\subsection{Control Architecture for Switching Between Variable Energy Levels and Contact with the External Environment}
\begin{figure}[t]
	\centering
	\includegraphics[width=0.9\columnwidth]{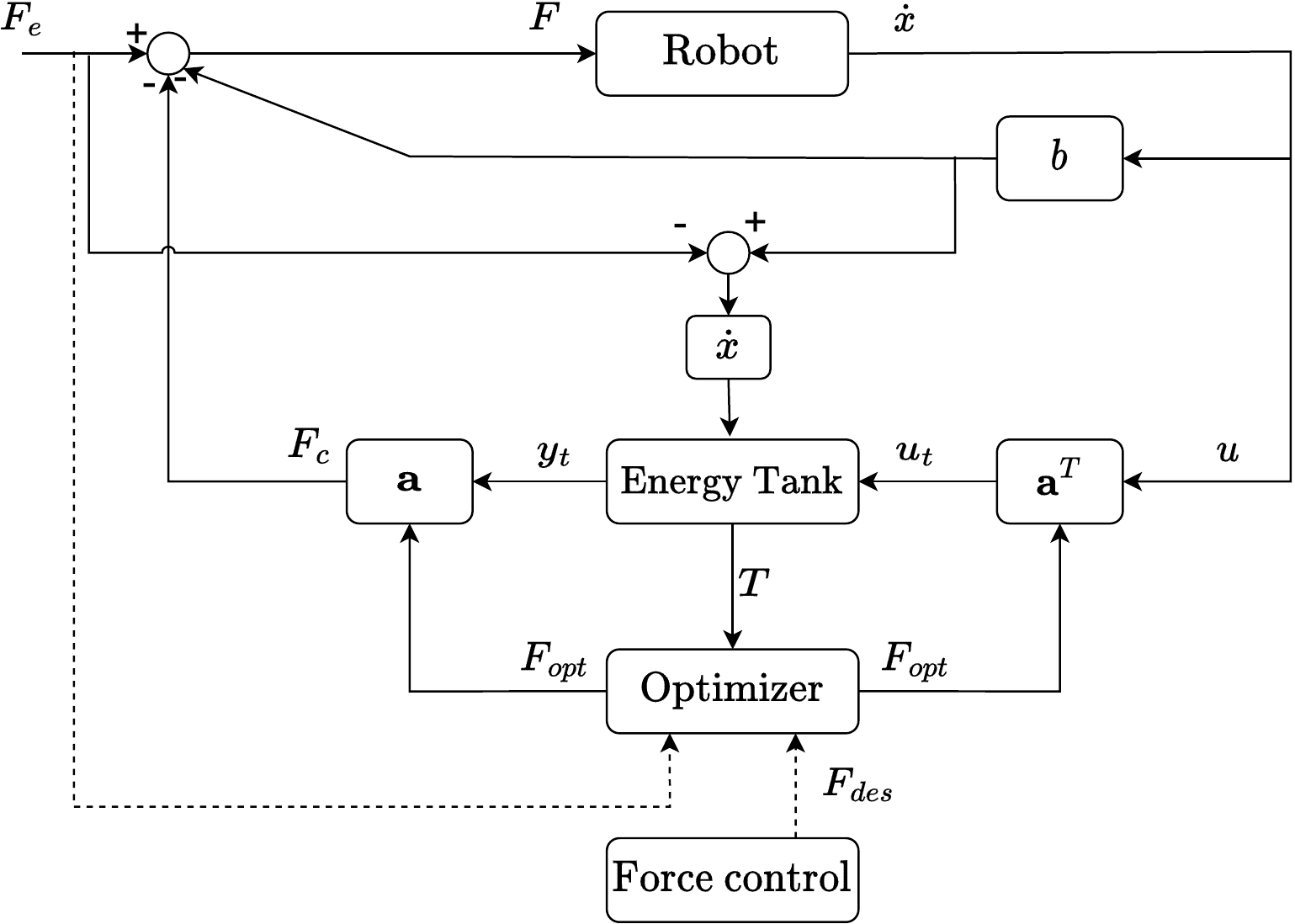}
	\caption{Modulated energy tank architecture with external energy flows. By means of energy re-routing, external power injections/extractions are properly handled, without affecting the bounding process.}
	\label{fig:ATA_ext}
\end{figure}

The control architecture presented in Sec.~\ref{sec:single} allows to bound the kinetic energy in free motion to a set value. On the other hand, when in contact with the external environment, the power injected through the port $(\mathbf{F}_{e}(t), \dot{\textbf{x}}(t))$, i.e. directly on the robot, can lead to uncontrolled variations of the kinetic energy.
Indeed, according to Prop. \ref{prop:ATAnoF}, the maximum allowable amount of energy for the robot is already stored in the closed loop system, since \mbox{$T(0) - \varepsilon +H(0)=\bar H$}. Thus, any uncontrolled additional energy injection on the robot (e.g., pushing the robot along its direction of motion) from external sources could violate the energy bound condition. On the contrary, uncontrolled external energy extractions (e.g., deceleration due to collisions) can degrade the overall performance, since the tank would not account for the energy loss. We can address this issue by properly re-routing the external energy in the tank, further augmenting the dynamics as follows:
\begin{equation}
  \label{eq:tankfe}
  { 
  \left \{
    \begin{array}[l]{l}
      \dot{x}_t(t)=\frac{1}{x_t(t)}\left[-\mathbf{F}_{e}(t)^T\dot{\textbf{x}}(t) + b(t)\dot{\textbf{x}}(t)^T \dot{\textbf{x}}(t) \right] +\mathbf{a}(t)^T\mathbf{u}(t)\\
      \mathbf{F}_{c}(t)=\mathbf{a}(t)y_t(t),
    \end{array}
  \right.}
\end{equation}
in which $b$ is a variable damper that we activate only for dissipating the external energy injection whenever this would violate the energetic bound, as shown in Fig. \ref{fig:ATA_ext}, i.e.,
\begin{equation}
  \label{eq:b}
b(t)=  \left\{
    \begin{array}[l]{ll}
      \frac{\mathbf{F}_{e}(t)^T\dot{\textbf{x}}(t)}{\dot{\textbf{x}}(t)^T\dot{\textbf{x}}(t)} & \mbox{if } \mathbf{F}_{e}(t)^T\dot{\textbf{x}}(t)  >0 \text{ and } T(t) = \varepsilon
      \\
      0 & \mbox{else.}
    \end{array}
    \right.
\end{equation}
This is an emergency term which dissipates the external injection only when $H(t) = \bar H$, and is thus rarely activated. The additional terms in \eqref{eq:tankfe} allow the tank to track the external energy flowing in/out of the robot. In case of injection, i.e., $\mathbf{F}_{e}(t)^T \dot{\mathbf{x}}>0$, the corresponding amount of power is extracted from the tank. Conversely, any external extraction, i.e. $\mathbf{F}_{e}(t)^T \dot{\mathbf{x}}<0$, is injected back into the tank.
\begin{prop}\label{prop:ATAF}
Consider the robot \eqref{eq: cart_dyn_model} interconnected with the tank \eqref{eq: modulated_tank}. If $T(0)=\bar H-H(0)+\varepsilon$ and \mbox{$T(t) \geq \varepsilon \,\, \forall t \geq 0$}, then $H(t)\leq \bar H \,\, \forall t\geq 0$, both in free motion and in contact with the external environment.
\end{prop}
\begin{pf}
Using \eqref{eq:tankfe} and \eqref{eq:b}, the following balance holds:
\begin{equation}
  \label{eq:ATAFcomp-balance}
  \begin{array}{ll}
     \dot{H}(t)+\dot{T}(t)&=  \\
     &= \mathbf{F}(t)^T\dot{\textbf{x}}(t)-b(t)\dot{\textbf{x}}(t)^T\dot{\textbf{x}}(t) 
     \\& - \left(\mathbf{F}_{e}(t) + \mathbf{F}_{c}(t)\right)^T\dot{\textbf{x}}(t) +b(t)\dot{\textbf{x}}(t)^T\dot{\textbf{x}}(t)=  \\
     &= 0,
  \end{array}
\end{equation}
Thus, by integrating, we get the following energy balance
\begin{equation}
  \label{eq:ATAFcomp-energy}
  H(t)\leq H(0)+T(0)-T(t)\leq H(0)+T(0)-\varepsilon,
\end{equation}
where the inequality comes, again, from the fact that \mbox{$T(t)\geq \varepsilon$ $\forall t\geq0$}. Therefore, setting \mbox{$ T(0) = \bar H -H(0) + \varepsilon$} we get from \eqref{eq:ATAFcomp-energy} that \mbox{$H(t)\leq \bar H$}, thus concluding the proof.
\end{pf}
Thus, the kinetic energy bound can be ensured and external injections/extraction of energy are accounted for via the interconnection in  \eqref{eq:tankfe}. Since none of the terms in \eqref{eq:tankfe} depends upon the control input, these can simply be added into the optimization problem \eqref{eq: opt_max_energy_freemotion}, preserving the convexity of the problem. In particular, we can utilize the architecture to comply directly with the energetic bounds imposed by PFL \eqref{eq:max_energy}, by setting $\bar H = E_{max}$.

Nonetheless, the bounds introduced by PFL can often change during the robot motion, as the value of $E_{max}$ in \eqref{eq:max_energy} depends on the currently closest body region to the robot.
Let us consider the previously introduced energy levels  \mbox{$0<H_1\leq H_2 \leq \cdots \leq H_N$}. Each energy bound $H_i$ can be ensured at different times by modifying online \eqref{eq: opt_max_energy_freemotion}
\begin{equation} \label{eq: opt_max_energy_final}
    \begin{aligned}
        & \underset{\alpha(k)}{\text{minimize}}
        & & ||\alpha\mathbf{F}_{des}(k) - \mathbf{F}_{des}(k)||^2 \\
        & \text{subject to}
        & & \tau \alpha\textbf{F}_{des}(k)^T\dot{\textbf{x}}(k) + \tau P_{ext} + T(k-1) \geq \varepsilon(k),
        \end{aligned}
\end{equation}
where the term \mbox{$P_{ext} = \mathbf{F}_{e}(t)^T\dot{\textbf{x}}(t) + b(t)\dot{\textbf{x}}(t)^T \dot{\textbf{x}}(t)$} encompasses the power flows due to the environmental interaction, and in which we vary online the lower bound $\varepsilon (k)>0$ in order to change how much energy is made available to the robot. Let $k_i$ be the discrete time instant at which the closest body part to the robot changes. Correspondingly, the bound on the kinetic energy of the robot switches to $H_i$. In order to address this switch, $\varepsilon (k_i)$ is changed s.t. \mbox{$\varepsilon(k_i) = T(0) - H_i + H(0)$}. In this way, the result of Prop.~\ref{prop:ATAF} keeps on holding despite the varying energy bound, thus guaranteeing stability also during the switching.

In this way, we have build an energy-based control architecture capable of limiting the kinetic energy of the robot to different arbitrary bounds, both in free motion and in interaction phase. Moreover, this can be performed without knowledge of the dynamic parameters (i.e., the inertial and Coriolis terms) of the robot in \eqref{eq: robot_dyn_model}.

\section{Simulations}\label{sec: simulations}
Simulations have been performed in order to validate the control architecture, using a KUKA LWR 4+ 7-DOF force-controlled manipulator in MATLAB, modeled as in \eqref{eq: robot_dyn_model}, whose dynamic behavior is computed using the Robotics Toolbox \cite{corke1996robotics} with a sampling time of 1$ms$.
The robot is set to accomplish a simple motion task between two poses. The desired input $ F_{des}$ is generated via a standard PD controller, for the sake of generality.

First, following the procedure in Sec.~\ref{sec:background}, we compute $v_{max}$ according to the TS. During the first part of the motion, the closest human body part is the chest. From the tables in \cite{ISO_TS_15066:2016}, we retrieve the maximum force for this part ($140N$), together with its stiffness ($25N/mm$), mass ($40kg$), and maximum energy ($E_{max1} = 1.6J$), under the assumption that the robot is carrying no payload. Using \eqref{eq:vISO}, the maximum velocity for the first motion according to the TS would be $v_{max1} = 0.29m/s$. 

In order to directly comply with the energetic bound of the TS instead, we leverage our architecture in Sec.~\ref{sec: bounds} and approximate the control input $F_{des}$ by means of the optimization problem in \eqref{eq: opt_max_energy_final}. Since the robot is in free motion, $P_{ext} = 0W$ in this case. We initially set the tank lower bound to a conservative value of $\varepsilon_1 = 3.4J$. Then, as previously described, we initialize the tank as \mbox{$T(0) = \varepsilon_1 + E_{max1} = 5.0J$}. In this way, assuming $H(0) = 0J$ (the robot is initially stopped), we can limit the kinetic energy of the robot s.t. $H(t) \leq E_{max1}$. The scaled input $F_c = \alpha F_{des}$ is then applied to the robot.

During the second part of the motion, the threatened body part changes to the shoulders of the operator. Performing the previous procedure according to the TS for this body part, we obtain the new values for the maximum velocity $v_{max2} = 0.37 m/s$ and transferred energy $E_{max2} = 2.5J$.
This switch can be managed in the architecture by simply changing the lower bound of the tank to $\varepsilon_2 = T(0) - E_{max2} = 2.5J$, thus allowing additional task energy.

\begin{figure}[!th]
	\centering
	\includegraphics[width=0.8\columnwidth]{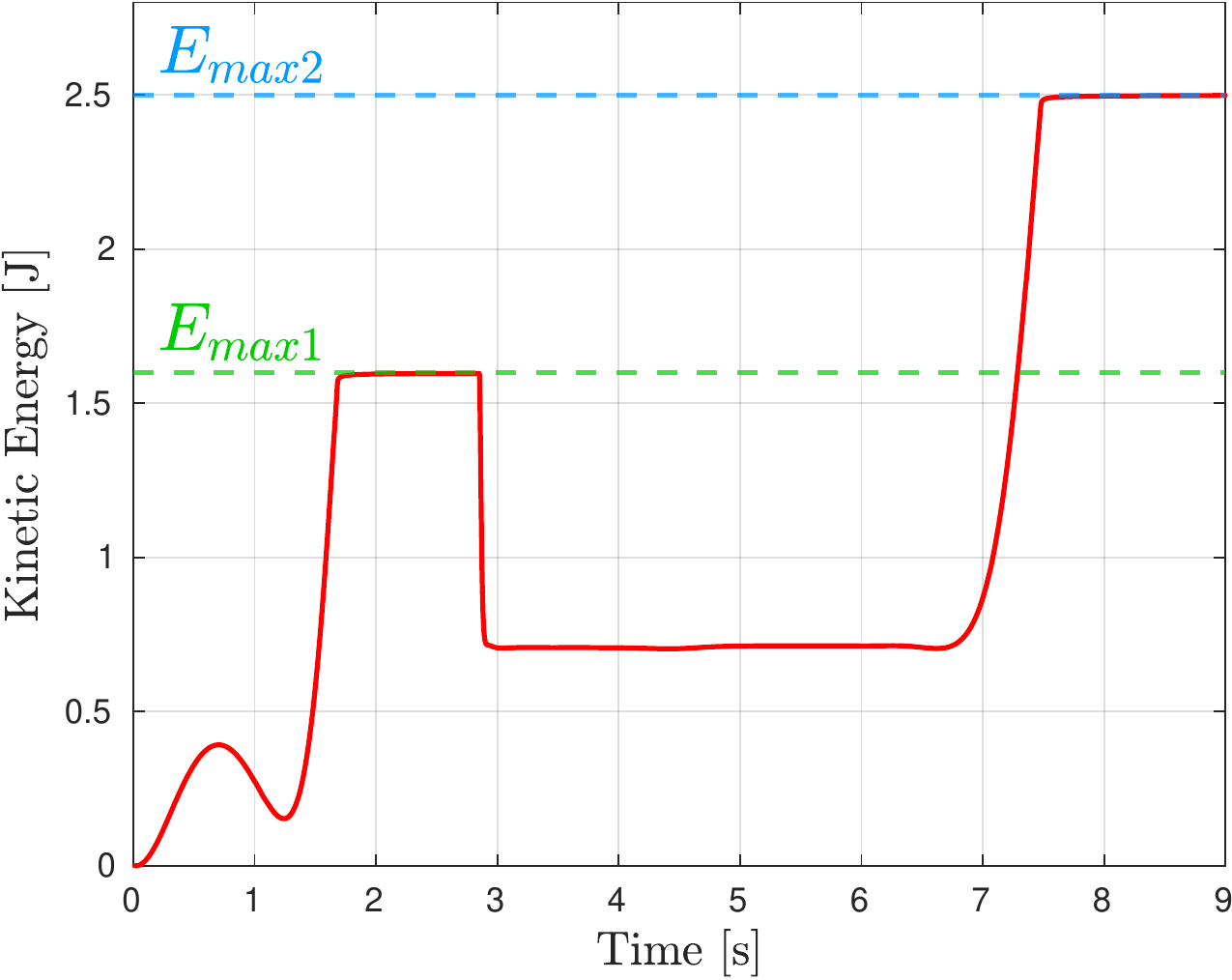}
	\caption{Evolution of the kinetic energy of the robot during task execution. }
	\label{fig:kin_energy_evo}
\end{figure}

\begin{figure}[!th]
	\centering
	\includegraphics[width=0.8\columnwidth]{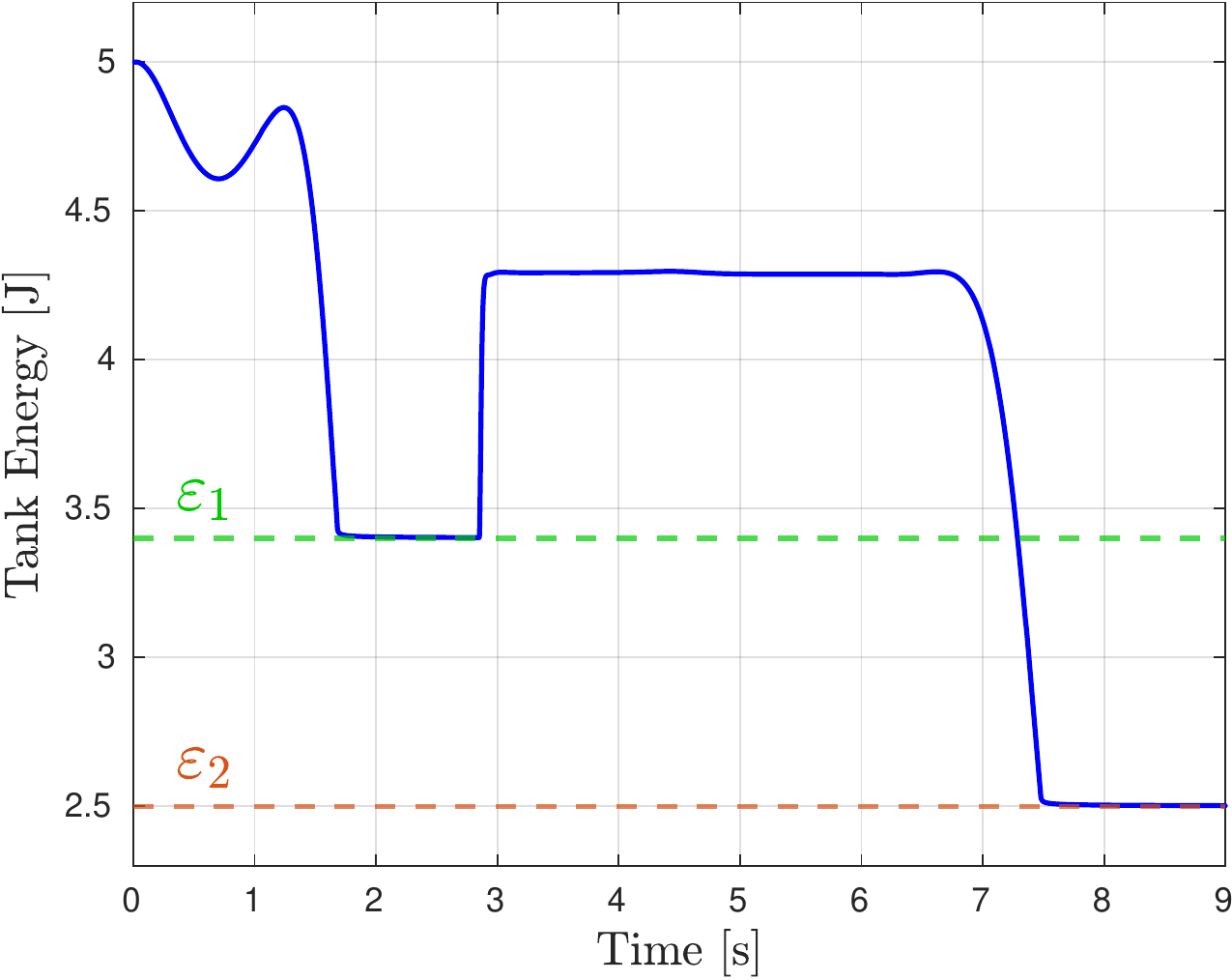}
	\caption{Evolution of the energy in the tank during task execution. }
	\label{fig:tank_energy_evo}
\end{figure}


\begin{figure}[!th]
	\centering
	\includegraphics[width=0.8\columnwidth]{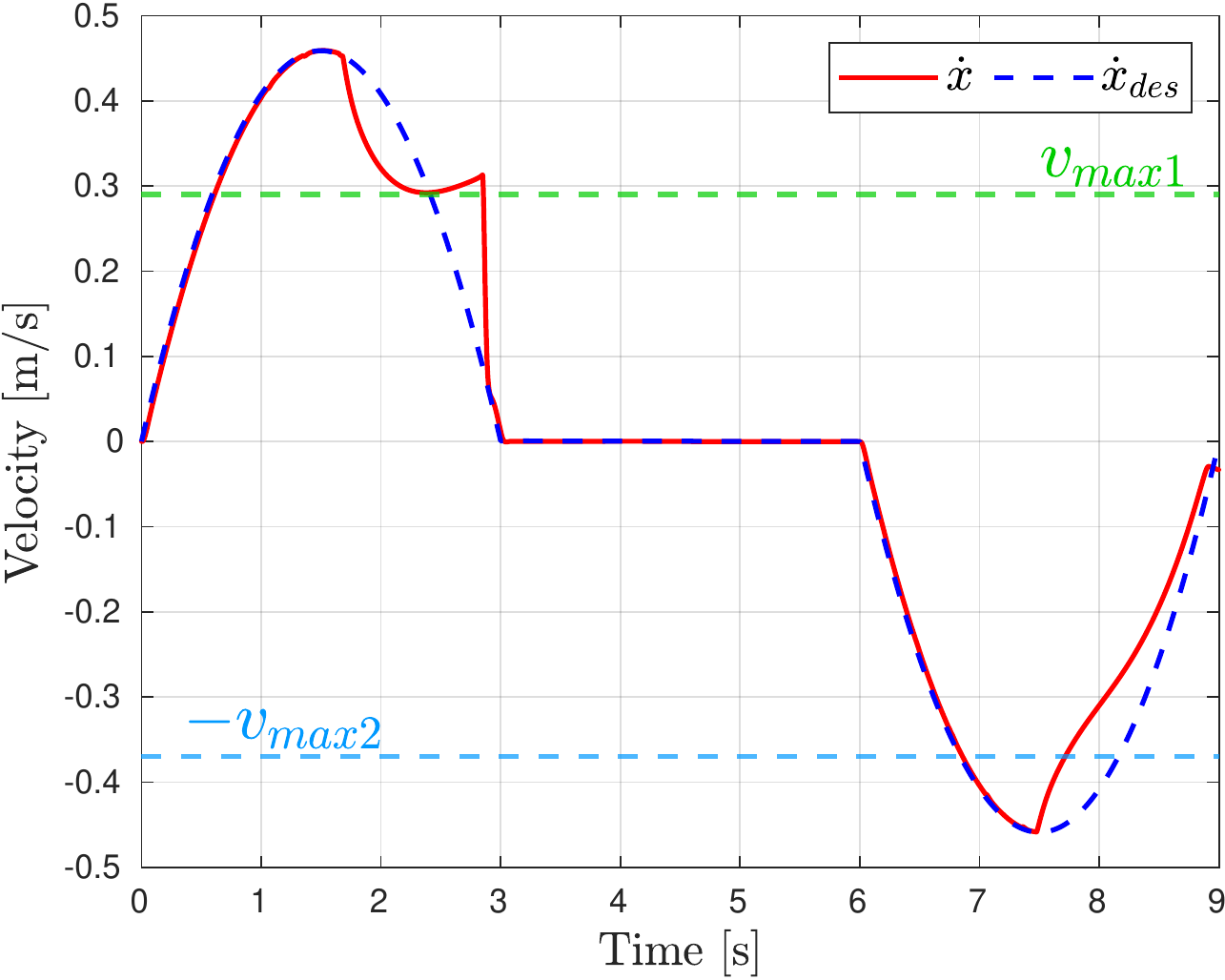}
	\caption{Velocity along the direction of motion during task execution. }
	\label{fig:vel}
\end{figure}

The results are showcased in Fig.~\ref{fig:kin_energy_evo}, Fig.~\ref{fig:tank_energy_evo} and Fig.~\ref{fig:vel}: during the first part of the motion, the robot follows the desired velocity profile as long as its kinetic energy remains below the safety bound $E_{max1}$, i.e., as long as the energy in the tank remains above $\varepsilon_1$. Whenever the bound is reached, the input is approximated via \eqref{eq: opt_max_energy_final}, trading off performance in order to guarantee safety. From Fig.~\ref{fig:vel} in particular, two important conclusions can be drawn. Firstly, that the bound on the robot velocity is generally conservative, as during both parts of the motion task we manage to move at an higher speed value than the one imposed by $v_{max1}$ and $v_{max2}$, while still complying with the energetic requirement of PFL. Secondly, that a velocity limit is not enough to guarantee the safety of the operator. Notice how, during the second part of the motion, the robot slows down after $7.5sec$ to a value lower than $v_{max2}$. This is because, due to postural conditions and to its self-motions, the kinetic energy of the robot is already at the maximum allowed value $E_{max2}$. Since our architecture is energy-based, we can easily take these factors into account during the input computations. It is easy to see that, in this case, the direct application of the velocity bound would instead lead to an unsafe behavior, i.e., the energetic bound $E_{max2}$ would not be respected. 

\section{Conclusions and Future Works}\label{sec: conclusion}
In this paper we developed a model-agnostic energy tank-based control architecture capable of complying directly with the time-varying energetic bounds of ISO TS 15066. The architecture is general purpose, and can be directly employed as a safety layer for any force/torque controlled robot. Future works aim at experimentally validating the architecture in a real collaborative scenario, providing a formal proof of stability and safety during bound-switching phases, and leveraging kinematic redundancy for reducing the apparent mass of the robot during the motion.
 
\bibliography{ifacconf}
\end{document}